\documentclass[runningheads]{llncs}
\usepackage{graphicx}
\usepackage{amsmath}
\usepackage{amssymb}
\usepackage[colorlinks=true]{hyperref}
\usepackage{booktabs} %
\usepackage{multirow} 
\usepackage[normalem]{ulem}
\useunder{\uline}{\ul}{}
\usepackage[ruled, linesnumbered]{algorithm2e}
\begin{document}
\title{Multiple Instance Learning with Mixed Supervision in Gleason Grading}
\titlerunning{Multiple Instance Learning with Mixed Supervision in Gleason Grading}
%
\author{Hao Bian\inst{1}\orcidID{0000-0002-3809-9574} \and
Zhuchen Shao\inst{1}\orcidID{0000-0002-9554-8831} \and
Yang Chen\inst{1}\orcidID{0000-0001-6410-0922}\and
Yifeng Wang\inst{2}\orcidID{0000-0001-8317-8434}\and
Haoqian Wang\inst{1}\orcidID{0000-0003-2792-8469}\and
Jian Zhang\inst{3}\orcidID{0000-0001-5486-3125}\and
Yongbing Zhang\thanks{Corresponding author. \\Co-first authors: Hao Bian, Zhuchen Shao, Yang Chen.}\inst{2}\orcidID{0000-0003-3320-2904}}

%
\authorrunning{H. Bian et al.}
%
\institute{Tsinghua Shenzhen International Graduate School, Tsinghua University \\
	\email{bianh21@mails.tsinghua.edu.cn}\and
Harbin Institute of Technology (Shenzhen)\\
\email{ybzhang08@hit.edu.cn}\and
School of Electronic and Computer Engineering, Peking University
}
%
\maketitle              
\begin{abstract}
With the development of computational pathology, deep learning methods for Gleason grading through whole slide images (WSIs) have excellent prospects. Since the size of WSIs is extremely large, the image label usually contains only slide-level label or limited pixel-level labels. The current mainstream approach adopts multi-instance learning to predict Gleason grades. However, some methods only considering the slide-level label ignore the limited pixel-level labels containing rich local information. Furthermore, the method of additionally considering the pixel-level labels ignores the inaccuracy of pixel-level labels. To address these problems, we propose a mixed supervision Transformer based on the multiple instance learning framework. The model utilizes both slide-level label and instance-level labels to achieve more accurate Gleason grading at the slide level. The impact of inaccurate instance-level labels is further reduced by introducing an efficient random masking strategy in the mixed supervision training process. We achieve the state-of-the-art performance on the SICAPv2 dataset, and the visual analysis shows the accurate prediction results of instance level. The source code is available at \url{https://github.com/bianhao123/Mixed_supervision}.

\end{abstract}

\section{Introduction}
Prostate cancer is the second most common cancer in men, with a large number of new cases every year. For the diagnosis of prostate cancer, whole slide images (WSIs) are currently the gold standard for clinical diagnosis. Pathologists analyze WSIs by visual inspection, classify tissue regions, detect the presence of one or more Gleason patterns, and ultimately make a diagnosis based on a composite Gleason score. For example, a composite grade of 5 + 4 = 9 would be assigned to a sample where the primary Gleason grade is 5 and the secondary is 4. However, pathologists still face many challenges in Gleason grading: (1) Since WSIs are of enormous data volume,  observation and analysis are time-consuming; (2) WSIs are of poor quality, with artifacts and tissue folding. Therefore, some machine learning and deep learning algorithms provide automatic solutions for Gleason grading. However, due to the long labeling time and the need for professional medical knowledge in Gleason grading, WSIs usually only contain slide-level labels or some limited pixel-level labels. In addition, the refined pixel-level labels may be overlapping and inaccurate sometimes.

When only slide-level labels are available, some weakly supervised multiple instance learning (MIL) algorithms are proposed to predict the slide-level labels of WSIs automatically. At present, commonly embedding-based MIL methods can be divided into two categories: attention-based MIL methods \cite{ilse2018attention,lu2021data,shi2020loss},  and correlated MIL methods \cite{shao2021transmil,myronenko2021accounting,li2021dual}. The attention-based MIL method is mainly based on the bypass attention mechanism, which provides additional contribution information for each instance through learnable attention weight. Correlated MIL method mainly includes non-local attention mechanism and self-attention mechanism. These methods can capture the dependencies between instances by calculating the attention scores.

When both slide-level label and limited pixel-level labels are available, methods such as \cite{anklin2021learning,tourniaire2021attention} are proposed to deal with mixed supervision scenarios, which can promote the classification performance. However, these mixed supervision methods do not consider the impact of limited inaccurate pixel-level labels on model performance.

In this work, we propose a mixed supervision Transformer based on the MIL framework. First, pixel-level labels are converted into instance-level labels through the superpixel-based instance feature and label generation. The slide-level multi-label classification task and the instance-level multi-classification task are jointly trained in the training process. Second, we adopt an effective random masking strategy to avoid the performance loss caused by the inaccurate instance-level labels. At the same time, we perform 2D sinusoidal position encoding on the spatial information of the instance, which is beneficial for the correlation learning between instances. Our method achieves the best slide-level classification performance, and the visual analysis shows the instance-level accuracy of the model in Gleason pattern prediction.

\section{Method}

\subsection{Problem Fomulation}
Gleason grading is a  multi-label MIL classification task. A WSI is regarded as a bag $X$, which contains $N$ instances $\left\{{x}_{1}, {x}_{2}, \ldots, {x}_{N}\right\}$ and each instance represents a pixel set with a proper size. The instance-level labels  $ \left\{ y_1, y_2, \ldots, y_N \right\}$ are unknown, and the bag-level label $Y$ is a ground truth set of $\ell$ binary labels $\left\{p_{1}, p_{2}, \ldots, p_{\ell}\right\}, p_{i} \in\{0,1\}$. 

\begin{figure}
	\centering
	\includegraphics[width=1\linewidth]{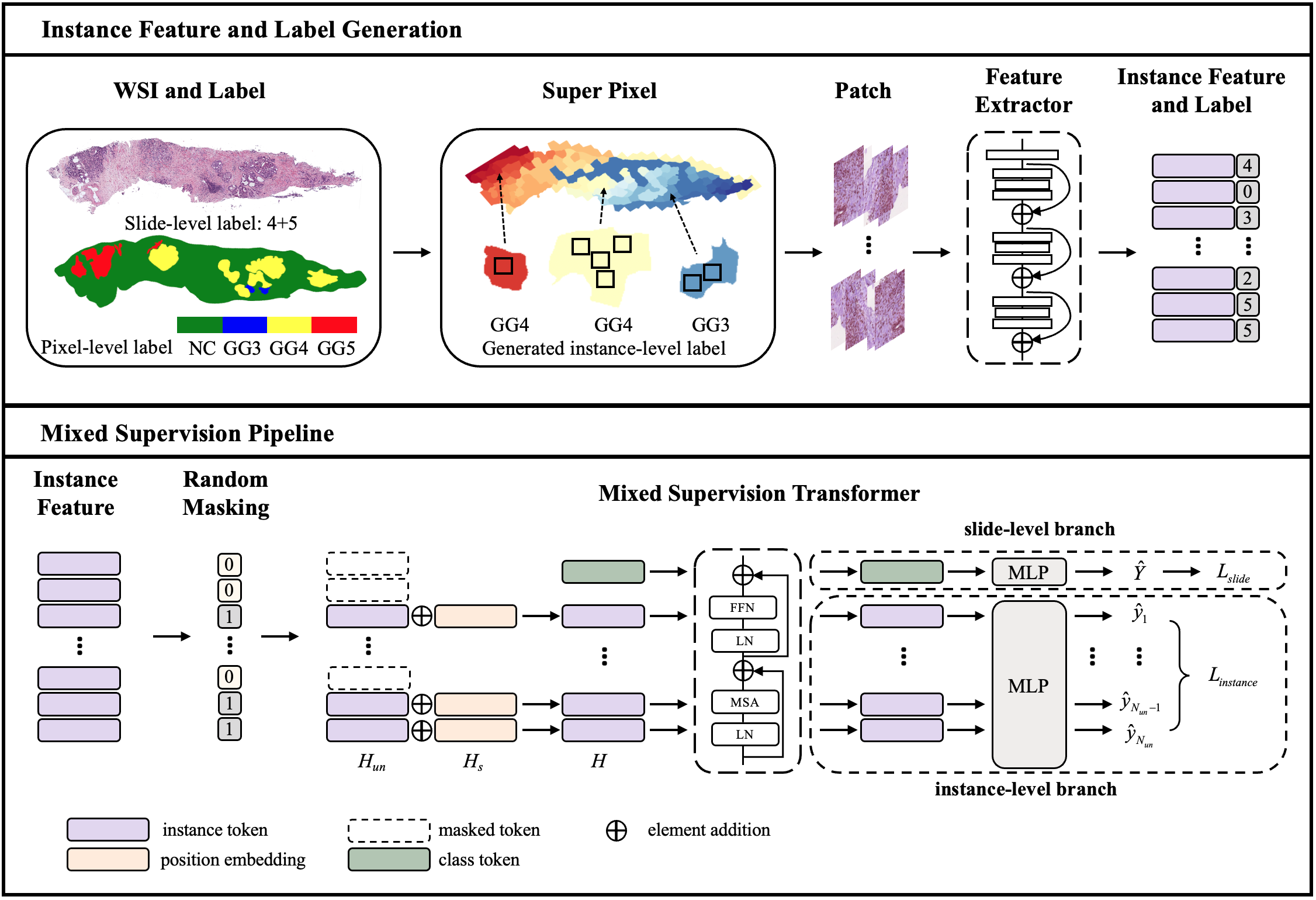}
	\caption{Overview of our proposed method. In the first step, we obtain instance-level features and labels according to the generated superpixel regions. In the second step, we adopt a random masking strategy to train a mixed supervision Transformer, which utilizes both slide-level label and instance-level labels.
	}
	\label{fig:overview}
\end{figure}

In practice, besides slide-level labels, Gleason grading task also has limited pixel-level labels sometimes. Therefore, based on mixed supervision, it would be beneficial to improve the accuracy of Gleason grading by effectively utilizing the two types of labels.
However, pixel-level labels may be inaccurate. 
Therefore, the mixed supervision of the Gleason grading can be divided into two steps, as shown in Fig. \ref{fig:overview}. First, the inaccurate pixel-level labels are employed to get more reliable instance-level labels. Next, both some instance-level labels and the slide-level label are utilized for mixed supervision model training. In the following, we will provide the description of two steps in detail.

\subsection{Instance Feature and Label  Generation}

Although pixel-level labels in Gleason grading are not always accurate, we can convert the inaccurate pixel-level labels into more reliable instance-level labels. However, patch-based methods cannot obtain a reliable instance-level label containing same tissue structures within one rectangular patch.
Inspired by the method in \cite{anklin2021learning}, we first filter the blank area and use stain normalization on different WSIs. Then we employ the simple linear iterative clustering  (SLIC) algorithm \cite{achanta2010slic} to extract superpixel tissue regions. The area of each superpixel can be considered as an instance.  Since the superpixel region is generated according to the similarity of tissue texture structure, each superpixel region contains most of the same tissues and has a smoother boundary than the rectangular block. Therefore, we assign the pixel-level labels with the largest proportion as the labels of instance-level.

Considering the irregularities and different sizes of the generated superpixel regions, we extract the instance feature as follows.
First, based on the centroid of each superpixel region, we cut it into one or more patches sized of $224\times 224$. Then, we employ ImageNet pre-trained mobilenetv2 to extract $d$ dimensional features ($d$ is 1280). It is worth noting that in the case of cutting out multiple patches, we average the feature of each patch as the instance-level feature corresponding to each superpixel region.

\subsection{Mixed Supervision Pipeline}
This section introduces the training pipeline of mixed supervision. Firstly, a random masking strategy  is employed to generate the unmasking instance tokens.
Then, a mixed supervision Transformer is designed by utilizing both slide-level label and instance-level labels to achieve more accurate Gleason grading at slide level.

\subsubsection{Radom Masking Strategy}
For Gleason grading, pixel-level labels may be inaccurate, which will cause the error in generated instance-level labels and poor performance of mixed supervision. 
To assist the training of mixed-supervised network, we adopted an effective sampling strategy (random masking) to optimize the training process, inspired by MAE \cite{he2022masked}. In each training epoch, we sampled the instance token and corresponding label without replacement according to the uniform distribution. Therefore, unmasked instance tokens $H_{un}=\{  z_1, \ldots, z_{N_{un}}\}$  are obtained, where $N_{un}=(1-m) \times N, z_i \in R^d$ and $m$ is masking ratio. The uniform distribution sampling ensures that the unmasked instance tokens are distributed in the entire WSI area, enabling the mixed supervision Transformer to encode the information of the entire WSI as much as possible. The advantages of random masking strategy are two folds: (1) Reducing the impact of label inaccuracy. (2) Reducing computation and memory occupation.

\subsubsection{Mixed Supervision Transformer}

The Gleason grading task has slide-level and instance-level supervision information. The Transformer structure contains two types of token: class token and instance token, which correspond exactly to the two types of supervision information in the Gleason grading. Based on this property, we design the mixed supervision Transformer.
The whole structure can be divided into two branches: slide-level branch and instance-level branch, corresponding to class token output and instance token output,  respectively.

To obtain class token output and instance token ouput, firstly, we encode the spatial information of unmasking instance tokens. We obtain the centroid coordinates $\left( p_x, p_y \right)$ of the superpixel area corresponding to each instance and encode coordinate information  by 2D sinusoidal position encoding \cite{vaswani2017attention}. Then, similar to  \cite{dosovitskiy2021an,cai2022mask,cai2022mst++},   we add a learnable class token. The class token and all the instance tokens are combined and fed into Transformer, which can capture slide-level and instance-level information, respectively. The detailed calculation of class token and instance token is shown in Algorithm \ref{algorithm1}.

\begin{algorithm}[t]
	
	\label{algorithm1}
	\caption{Class Token and Instance Token  Calculation}
	\KwIn{unmasking tokens $H_{un}=\{  z_1, \ldots, z_{N_{un}}\},  z_i \in R^d, H_{un} \in R^{N_{un}\times d}$}
	\KwOut{class token outputs $\tilde{Y}$, instance token output $\tilde{y_1}, \ldots, \tilde{y}_{N_{un}}$}
	\tcp{1. add 2D spatial position encoding to unmasking token $H_{un}$}
	$\left( p_{i, h}, p_{i, w} \right) \leftarrow$ centroid coordinate of each instance  $z_i$ \;
	
	$PE_{(i, pos, 2 j)}=\sin \left(\frac{pos}{10000^{2 j / d_{ {half }}}}  \right)$,
	$PE_{(i, pos, 2 j+1)}=\cos \left(\frac{pos}{10000^{2 j / d_{{half }}}} \right) $\quad\quad\quad\quad\quad\quad\quad\quad
	$\triangleright$ $ pos \in \{\frac{p_{i, h}}{100}, \frac{p_{i, w}}{100}\}, d_{half} = \frac{d}{2}, j \in [0, d_{half} - 1]$\;
	
	$s_i  \in R^d \leftarrow \operatorname{CONCAT}[PE_{i, h}, PE_{i, w}]$ \quad $\triangleright$ encoding $p_{i,h}$ and $p_{i, w}$, and concatenate two-dimensional embeddings\;
	
	$h_i \leftarrow z_i + w s_i $   \quad $\triangleright$ add instance token and spatial postion token, $w$ is 0.1\;

	\tcp{2. correlation learning between instances by Transformer}
	$h_{class} \in R^d$ $\leftarrow$  set a learnable class token \;
	$H^{(0)} = \{h_{class},  h_1, \ldots, h_{N_{un}}\} \in R^{(N_{un} + 1) \times d}$ \quad $\triangleright$ concatenate class token and instance token \;
	
	\For{$l \in [0: 1: L-1]$}{$H^{(l+1)} = \operatorname{Transformer}(H^{(l)})$}

\tcp{3. class token output and instance token output}
$\tilde{Y}\leftarrow H^{(L)}[0]$ \quad$\triangleright$ input to slide-level branch\;
$\tilde{y_1}, \ldots, \tilde{y}_{N_{un}} \leftarrow H^{(L)}[1], \ldots, H^{(L)}[N_{un}] $ \quad$\triangleright$  input to instance-level branch\;

\end{algorithm}

The slide-level branch is actually  a multi-label classfication task. We use multi-layer perception to predict $\hat{Y}$ for the class token output $\tilde{Y}$.
Through the sigmoid layer, the slide-level loss $L_{bag}$ is calculated via a multi-label weighted cross entropy loss function $L_1$ with slide-level label $Y$: 
\begin{equation}
L_{slide}= L_{1}(Y, sigmoid(\hat{Y})).
\end{equation}

For the instance-level branch, since we generate an instance-level label for each superpixel instance, it is regarded as a multi-category task. So we use multi-layer perception to predict $\hat{y_{i}}$ for the  instance token output $\tilde{y_{i}}$. Through the softmax layer, the instance-level loss  $L_{instance}$ is calculated by a multi-category weighted cross entropy loss function $L_2$ with the instance-level label $y_i$: 

\begin{equation}
L_{instance}=L_{2}(y_i, softmax(\hat{y_i})).
\end{equation}
To optimize the model parameters, we minimize the following loss function:
\begin{equation}
L_{total} = \lambda L_{slide} +(1-\lambda )\sum_{k} L_{ {instance }},
\end{equation}
where $\lambda \in [0, 1]$, and $\lambda$ is set to 0.5 in our experiment. 

\section{Experiments}
\subsubsection{Dataset}
We evaluate our method on the SICAPv2 dataset \cite{silva2020going} for the Gleason grading task. SICAPv2 is a public collection of prostate H\&E biopsies containing slide-level labels (i.e., Gleason scores for each slide) and pixel-level annotations (18783 patches of size $512 \times 512$). SICAPv2 database includes 155 slides from 95 different patients. The tissue samples are sliced, stained, and digitized by the Ventana iScan Coreo scanner at $40\times$ magnification. Then the WSIs are obtained by downsampling to $10\times$ resolution, and Gleason's total score is assigned for each slide tissue. The main Gleason grade (GG) distribution in each slide is as follows: 36 noncancerous areas, 40 samples are Gleason grade 3, 64 samples are Gleason grade 4, and 15 samples are Gleason grade 5 (NC, GG3, GG4, and GG5). We randomly split the patient data in the ratio of training: validation: test = 60: 15: 25 and use 4-fold cross-validation for all experiments. Due to unbalanced data distribution, we use the StratifiedKFold method to ensure similar label ratios on the training, validation, and test sets. 

\subsubsection{Implementations}
We implement our method in PyTorch-Lightning and train it on a single NVIDIA GeForce RTX 3090 24 GB GPU.
In the mixed supervision Transformer, we employ 2 stacked Transformer blocks with 6 heads, and other configurations are similar to \cite{jiang2021all}. For 2D sinusoidal position encoding, we set the maximum $pos$  as 200. And the embedded dimension is set to 1280 as the instance feature dimension.
For the training process, the batch size is 1, and the grad accumulation step is 8. The Ranger optimizer \cite{ref_ranger} is employed with a learning rate of 2e-4 and weight decay of 1e-5. The validation loss is used as the monitor metric, and the early stopping strategy is adopted, with the patience of 20. We use macro AUC as the evaluation metric.

\subsubsection{Baselines} 
We compared our method with attention based methods such as ABMIL \cite{ilse2018attention}, CLAM \cite{lu2021data}, Loss-Attention \cite{shi2020loss}, corelated based methods such as DSMIL \cite{li2021dual}, AttnTrans \cite{myronenko2021accounting}, TransMIL \cite{shao2021transmil},  and GNN based method SegGini \cite{anklin2021learning}. 
In our experiment, we reproduce the baselines' code in the Pytorch-Lightning framework based on the existing code. The data processing flow of SegGini is consistent with our method. 
Other methods follow the CLAM standard processing step to extract patch features with the patch size of 224. And their parameters refer to the default parameter template used for segmenting biopsy slides.

\subsubsection{Result and Discussion}
\begin{table}[]
	\centering
	\caption{Evaluations results on SICAPv2 dataset as $Mean\pm std$. The bold font is the best score and the underline is the second score.}
	\label{tab:mytable1}
	{%
		\begin{tabular}{c|c|cc}
			\hline
			supervision                                                                         & Method                              & AUC &  \\ \hline
			\multirow{6}{*}{\begin{tabular}[c]{@{}c@{}}slide-level \\ supervision\end{tabular}} & ABMIL\cite{ilse2018attention}       & $0.6574\pm0.0825$       &               \\
			& CLAM\cite{lu2021data}               & $0.6096\pm0.0783$       &               \\
			& DSMIL\cite{li2021dual}               & $0.5920\pm0.0656$       &               \\
			& LossAttn\cite{shi2020loss}          & $0.5778\pm0.0484$       &               \\
			& ATMIL\cite{myronenko2021accounting} & $\underline{0.9373\pm0.0294}$       &               \\
			& TransMIL\cite{shao2021transmil}     & $0.9152\pm0.0314$       &               \\ \hline
			\multirow{2}{*}{\begin{tabular}[c]{@{}c@{}}Mixed \\ supervision\end{tabular}}       & SegGini\cite{anklin2021learning}    & $0.7941\pm0.1011$       &               \\
			& Ours                                & $\mathbf{0.9429\pm0.0094}$       &               \\ \hline
		\end{tabular}%
	}
\end{table}

According to Table \ref{tab:mytable1}, the AUCs of some current SOTA methods, such as ABMIL, CLAM, DSMIL, LossAttn, are ranged from 0.5778 to 0.6574, which is far from satisfaction. The main reason is that Gleason grading is a multi-label task, each instance has different categories, and the correlation between instances should be considered when classifying. The above methods are based on bypass attention, and the model scale is too small to efficiently fit the data, so the performance is relatively poor.
ATMIL and TransMIL models are Transformer-based models, which mainly adopt the multi-head self-attention mechanism. These models both consider the correlation between different instances and achieve better performance. However, the network structure of above methods does not utilize the instance-level labels, causing the AUC to be lower than our method from 0.0056 to 0.0277.
GNN based method SegGini is also a mixed supervision method, but it adopts all the instance-level label, which will be seriously affected by inaccurate labels.
The model we propose 
employs the random masking strategy and integrates the spatial position information of the instances in WSIs into the Transformer learning process to achieve the performance of SOTA (0.9429).

\begin{table}[]
	\centering
	\caption{Effects of masking instance token ration and spatial position encoding.}
	\label{tab:mytable2}
	\resizebox{\textwidth}{!}{%
		\begin{tabular}{c|c|c}
			\hline
			& w/o spatial position encoding & w/ spatial position encoding \\ \hline
			masking $0\%$          & ${0.9273\pm0.0103}$             & $0.9267\pm0.0148$        \\
			masking $10\%$          & $0.9337\pm0.0118$             & ${0.9405\pm0.021}$        \\
			masking $25\%$          & $0.9247\pm0.0153$             & ${0.9415\pm0.0163}$                                \\
			masking $50\%$            & $\mathbf{0.9339\pm0.0210}$            & $\mathbf{0.9429\pm0.0094}$                               \\ 
			only slide label     & $0.9172\pm0.0145$            & ${0.9190\pm0.0201}$                                 \\ \hline
		\end{tabular}%
	}
\end{table}

\begin{figure}
	\centering
	\includegraphics[width=1.0\linewidth]{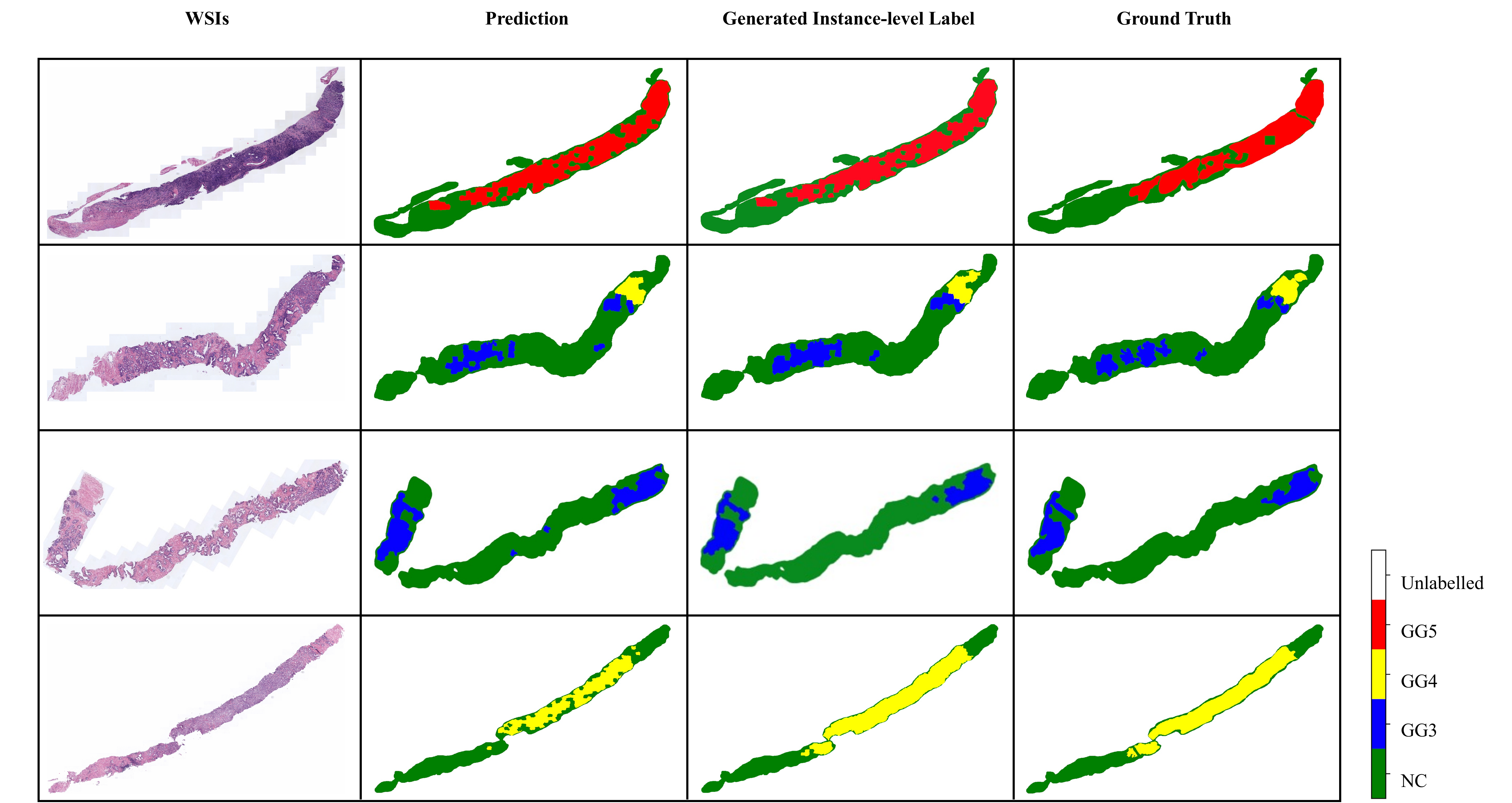}
	\caption{Gleason pattern prediction visualization.}
	\label{fig:vis}
\end{figure}

According to Table \ref{tab:mytable2}, we have the following observations:  (1) The performance of the model using slide-level label alone is not better than other models with mixed supervision. It indicates that adding the instance-level label to each instance token in the Transformer model can improve the slide-level classification. (2) When random masking ratios are 10\%, 25\% and 50\%, the model's performance is about 0.0160 better than using full instance token labels (masking 0\%), which shows that the strategy of random masking is effective. (3) The spatial position encoding can improve the performance in most experiment settings.
\subsubsection{Visual Analysis}
The motivation of our mixed supervision Transformer is that the class token corresponds to the slide information, and the instance token corresponds to the local superpixel information. The combination of these two types of label can improve the utilization of supervision information. In Fig. \ref{fig:vis}, we show the Gleason pattern prediction of the instance-level branch. It can be seen that the label of each superpixel area can be predicted more accurately.

\section{Conclusion}

Gleason grading is a multi-label MIL classification task, which has slide-level labels and limited pixel-level labels sometimes. For this task, we propose a method composed of two steps: (1) instance feature and label generation; (2) mixed supervision Transformer. In the first step, we adopt the SILC algorithm to obtain more reliable instance-level labels from inaccurate pixel-level labels. In the second step, both instance-level labels and slide-level labels are utilized for training the mixed supervision Transformer model. Besides, we employ the random masking strategy to further reduce the impact of inaccurate labels. In the SICAPv2 dataset, we achieve state-of-the-art performance. Meanwhile, the visual analysis further shows that the instance-level branch can get more accurate pattern prediction. In the future, we will develop more interpretative masking strategies and optimize our model on larger datasets.

\subsubsection{Acknowledgements} 
\begin{sloppypar}
This work was supported in part by the National Natural Science Foundation of China (61922048 \& 62031023), in part by the Shenzhen Science and Technology Project (JCYJ20200109142808034 \& JCYJ20180508152042002), and in part by Guangdong Special Support (2019TX05X187).
\end{sloppypar}

%
%
%
 \bibliographystyle{splncs04}
 \bibliography{mybibliography}
 
\end{document}